\documentclass[final,5p,times,twocolumn,]{elsarticle}
\usepackage{amssymb}
\usepackage[table, dvipsnames]{xcolor}
\usepackage{booktabs}
\usepackage{graphicx}
\usepackage{soul}
\usepackage{amsmath}
\usepackage{algorithm}
\usepackage[noend]{algpseudocode}
\usepackage{tabularx}
\usepackage{natbib}
\usepackage{caption}
\usepackage{subcaption}
\usepackage{float}
\usepackage{multirow}
\usepackage{dblfloatfix}
\usepackage{hyperref}
\usepackage{pifont}
\usepackage{pgfplots}
\pgfplotsset{compat=1.17}
\usepackage[left]{lineno}  
\journal{Elsevier}

\begin{document}

\begin{frontmatter}

\title{APCoTTA: Continual Test-Time Adaptation for Semantic Segmentation of Airborne LiDAR Point Clouds}

\author[label1,label2,label3]{Yuan~Gao}
\author[label4]{Shaobo~Xia\corref{cor1}}
\ead{shaoboxia2020@gmail.com}
\author[label1,label2,label3]{Sheng~Nie}
\author[label1,label2,label3]{Cheng~Wang}
\author[label1,label2,label3]{Xiaohuan~Xi}
\author[label5]{Bisheng~Yang}

\cortext[cor1]{Corresponding author.}

\affiliation[label1]{organization={Aerospace Information Research Institute, Chinese Academy of Sciences}, 
	city={Beijing 100094},
	country={China}}
\affiliation[label2]{organization={International Research Center of Big Data for Sustainable Development Goals},
	city={Beijing 100094},
	country={China}}
\affiliation[label3]{organization={University of Chinese Academy of Sciences},
	city={Beijing 100094},
	country={China}}
   
\affiliation[label4]{organization={The Department
		of Geomatics Engineering, Changsha University of Science and Technology},
	city={Hunan 410004},
	country={China}}
	
\affiliation[label5]{organization={The State Key Laboratory of Information Engineering in Surveying, Mapping and Remote Sensing, Wuhan University},
	city={Wuhan 430079},
	country={China}}

\begin{abstract}
Airborne laser scanning (ALS) point cloud semantic segmentation is a fundamental task for large-scale 3D scene understanding. Fixed models deployed in real-world scenarios often suffer from performance degradation due to continuous domain shifts caused by environmental and sensor changes. Continuous Test-Time Adaptation (CTTA) enables adaptation to evolving unlabeled domains, but its application to ALS point clouds remains underexplored, hindered by the lack of benchmarks and the risks of catastrophic forgetting and error accumulation. To address these challenges, we propose APCoTTA (\textbf{A}LS \textbf{P}oint cloud \textbf{Co}ntinuous \textbf{T}est-\textbf{T}ime \textbf{A}daptation), a \textcolor{black}{novel} CTTA framework tailored for ALS point cloud semantic segmentation. APCoTTA consists of three key components. First, we adapt a gradient-driven layer selection mechanism for ALS point clouds, selectively updating low-confidence layers while freezing stable ones to preserve source knowledge and mitigate catastrophic forgetting. Second, an entropy-based consistency loss discards unreliable samples and enforces consistency regularization solely on reliable ones, effectively reducing error accumulation and improving adaptation stability. Third, a random parameter interpolation mechanism stochastically blends adapted parameters with source model parameters, further balancing target adaptation and source knowledge retention. Finally, we construct two benchmarks, ISPRSC and H3DC, to address the lack of CTTA benchmarks for ALS point cloud segmentation. Extensive experiments demonstrate that APCoTTA achieves superior performance on both benchmarks, improving mIoU by approximately 9\% and 14\% over direct inference. The new benchmarks and code are available at \url{https://github.com/Gaoyuan2/APCoTTA}. 

\end{abstract}


\end{frontmatter}

\section{Introduction}

Advances in helicopter and unmanned aerial vehicle (UAV) platforms have enabled flexible and efficient acquisition of airborne laser scanning (ALS) point clouds. These data provide critical spatial information for applications including mapping \cite{9459429,CHEN2025114957}, reconstruction \cite{song2020curved,qiao2022power}, and forest parameter estimation~\cite{luo2019estimating,LIU2022112844}. Semantic segmentation of ALS point clouds assigns semantic labels (e.g., terrain, vegetation, and buildings) to individual points, providing scene understanding and serving as a fundamental step for these downstream tasks.

ALS point cloud semantic segmentation models are often deployed in a fixed-parameter manner after training, which restricts their generalization to the source domain and closely related scenes. \textcolor{black}{This deployment paradigm conflicts with the inherent continuous spatiotemporal variability of ALS data in remote sensing. Spatially, as the flight platform moves, ALS point clouds are acquired in a streaming manner, resulting in progressively changing data distributions, for example, when the surveyed area gradually transitions from urban regions to rural landscapes. Temporally, repeated surveys of the same region further introduce persistent domain shifts, particularly in time-series observations across different seasons, where variations in vegetation phenology, illumination, and surface conditions continuously alter point cloud characteristics. Furthermore, weather and sensor-related factors further intensify these continuous shifts. Adverse weather conditions, such as strong sunlight, rain, or fog, can substantially affect return density and increase noise levels, leading to unstable point cloud characteristics. Differences in LiDAR sensor configurations and flight settings, including emission frequency, scanning patterns, and flight altitude, introduce additional variability in point density and geometric accuracy. Over time, gradual sensor degradation caused by mechanical vibrations and hardware aging further alters measurement stability and data quality~\cite{Dong2023BenchmarkingRO}. Together, these factors create non-stationary and progressively evolving distributions, posing significant challenges to fixed-parameter models in practical semantic segmentation of ALS point clouds.}

To address evolving domain distribution shifts, existing studies primarily adopt three strategies: model retraining, Unsupervised Domain Adaptation (UDA), and Test-Time Adaptation (TTA). However, model retraining necessitates high-quality, dense 3D annotated data \cite{GAO202672}, incurring prohibitive costs \cite{10677356}. While UDA methods alleviate the demand for target domain annotations, they still necessitate access to source domain data during the adaptation process, which is often impractical in remote sensing applications constrained by data privacy \cite{GAO202672}. Although TTA circumvents the dependency on source domain data and expensive annotations, existing methods \cite{Wang2021TentFT,niu2022efficient} are typically predicated on the assumption that the ``target domain distribution remains static'', relying on self-supervised objectives such as entropy minimization \cite{Wang2021TentFT,niu2022efficient} or pseudo-labeling \cite{chen2022contrastive,goyal2022test} for optimization. In ALS scenarios, however, continuous spatiotemporal shifts, sensor heterogeneity, and variations in observation conditions render this assumption untenable, leading to a sharp decline in the reliability of pseudo-labels and inducing error accumulation \cite{Wang2022ContinualTD,Wang2023ContinualTD}. Furthermore, long-term continual adaptation to dynamic target distributions renders the model prone to a gradual loss of source domain knowledge, ultimately triggering the issue of catastrophic forgetting \cite{song2023ecotta}.

\textcolor{black}{These limitations highlight the need for Continual Test-Time Adaptation (CTTA) tailored to ALS point clouds, which aims to enable continuous adaptation to evolving target domains without access to source data. Existing CTTA approaches, such as mean-teacher training~\cite{10.5555/3294771.3294885}, augmentation-averaged predictions~\cite{lyu2024variational}, contrastive regularization~\cite{Dbler2022RobustMT}, source parameter restoration~\cite{Brahma2022APF}, and reliable sample selection \cite{Wang2023ContinualTD}, have demonstrated effectiveness in 2D image tasks~\cite{Wang2022ContinualTD,yuan2023robust,maharana2025palm}.}
\textcolor{black}{However, since ALS point clouds are inherently unstructured and subject to dynamic geometric and density variations, the key assumptions underlying CTTA methods designed for image-based tasks exhibit limited applicability to them. A more detailed analysis is provided in Section~\ref{sec:dis}. Furthermore, existing studies predominantly focus on autonomous driving scenarios~\cite{Cao2023MultiModalCT} and object-level point clouds~\cite{jiang2024pcotta}, while ALS point cloud data remains underexplored. At present, two fundamental challenges remain to be addressed:}

\begin{itemize} 
\item \textcolor{black}{Existing CTTA methods are primarily designed for 2D imagery and are ill-suited for ALS point clouds, where direct application often leads to unstable adaptation, error accumulation, and catastrophic forgetting.}

\item The absence of a standardized benchmark for ALS CTTA prevents rigorous evaluation and comparison of methods under continuous domain shifts.

\end{itemize}

\textcolor{black}{
In this paper, we address these challenges within the specific context of ALS point clouds by proposing APCoTTA, a novel CTTA framework.
To address catastrophic forgetting, we investigate the transferability of image-based parameter-efficient fine-tuning strategies to ALS point clouds and find that full-parameter updates are unsuitable. Inspired by the insights from~\cite{huang2021importance,maharana2025palm} that the magnitude of gradients relative to a uniform distribution quantifies the model's familiarity with data, we introduce a Dynamic Selection of Trainable Layers (DSTL) module specifically designed for point cloud networks, which freezes stable layers and updates only geometry-sensitive ones to effectively mitigate catastrophic forgetting.
To address error accumulation caused by incorrect pseudo-labels, we propose an Entropy-Based Consistency Loss (EBCL). Unlike generic consistency losses, EBCL explicitly filters out incorrect pseudo-labels generated from low-confidence samples, ensuring the model learns strictly from reliable samples. Despite these measures, prolonged adaptation may still lead to catastrophic forgetting. To counter this, we propose a Randomized Parameter Interpolation (RPI) mechanism. Instead of the abrupt `hard' resets used in prior works, it employs a `soft' parameter fusion strategy to provide a smoother regularization trajectory for continuous adaptation. Finally, we construct two robust benchmark datasets, ISPRSC and H3DC, based on the ISPRS \cite{rottensteiner2012isprs} and H3D \cite{kolle2021hessigheim} datasets, to address the lack of publicly available CTTA datasets for ALS point cloud semantic segmentation. To summarize, our contributions are fourfold:}
\begin{itemize}
     \item \textcolor{black}{We introduce two robustness benchmarks for airborne LiDAR: ISPRSC and H3DC (Sec. \ref{datasets}). These benchmarks simulate seven common corruption types and fill the gap in standardized datasets for evaluating ALS point cloud segmentation under adverse real-world conditions.}
    \item \textcolor{black}{We propose APCoTTA (Fig. \ref{fig:framework}), a novel CTTA framework specifically tailored for ALS point cloud segmentation (Sec. \ref{Third}). Unlike generic CTTA frameworks, it addresses the severe continuous distribution shifts and low-confidence samples, ensuring stable adaptation without requiring source data.}
   \item \textcolor{black}{We propose three synergistic modules, namely the Dynamic Selection of Trainable Layers module (Sec. \ref{DSTL}), the Entropy-Based Consistency Loss module (Sec. \ref{EBCL}), and the Randomized Parameter Interpolation mechanism (Sec. \ref{RPI}), to address the catastrophic forgetting and error accumulation challenges in ALS scenarios.}
    \item \textcolor{black}{Extensive evaluations demonstrate that APCoTTA achieves mIoU improvements of approximately 9\% and 14\% over direct inference on the ISPRSC and H3DC benchmarks, respectively (Sec. \ref{experiment}), validating its effectiveness in handling diverse and evolving domain shifts in ALS scenarios.}
\end{itemize}

\section{Related Works}\label{second}

\subsection{Point Cloud Unsupervised Domain Adaptation}
Unsupervised Domain Adaptation (UDA) aims to improve model performance on an unlabeled target domain by leveraging labeled data from a source domain, despite the presence of distribution shifts between the two domains. It includes adversarial learning, feature alignment, self-supervised learning, and pseudo-labeling methods.

Wu et al. \cite{wu2019squeezesegv2} generate diverse synthetic data using simulators such as GTA-V and employ Geodesic Correlation Alignment to align the feature distribution differences between the source and target domains.
Yi et al. \cite{yi2021complete} complete sparse LiDAR point clouds and optimize completion quality with local adversarial learning. Semantic segmentation is then performed on the normalized completed point clouds, with results projected back to the target domain.
Xiao et al. \cite{Xiao_Huang_Guan_Zhan_Lu_2022} design PCT (Point Cloud Translator), a point cloud transformation method that reduces domain shift between synthetic and real point clouds by addressing appearance and sparsity differences separately.
Rochan et al. \cite{rochan2022unsupervised} address cross-domain generalization in LiDAR semantic segmentation using self-supervised learning. 
Saltori et al. \cite{saltori2022cosmix} perform data augmentation at both local and global levels and mitigate domain shift by cross-mixing point clouds between dual-branch symmetric networks.
Li et al. \cite{Li2023AdversariallyMS} reduce domain gaps by applying adaptive masked noise to source point clouds and leveraging adversarial training to simulate real noise distributions.
Although these methods have achieved promising performance, they face challenges in cases where source domain data is inaccessible due to privacy or other constraints.

\subsection{Test-Time Adaptation}
Test-Time Adaptation (TTA) refers to a paradigm that optimizes a model online during inference using only unlabeled test data to adapt to target domain shifts without requiring access to source data. 
Compared to UDA, TTA operates without source data and enables real-time model adjustment, making it suitable for scenarios with privacy constraints or dynamic data distributions.
TTA has rapidly advanced in the 2D image domain, including classification \cite{Wang2021TentFT,niu2022efficient,yuan2024tea}, segmentation \cite{Zhang2021AuxAdaptSA,wang2023dynamically}, and object detection \cite{Kim2022EvTTATA,Vibashan2022TowardsOD,Veksler2023TestTA}.

For 3D point cloud semantic segmentation,
Saltori et al. \cite{saltori2022gipso} propose an online unsupervised domain adaptation method that does not require access to source data. It adaptively selects high-confidence pseudo-labels through self-training and propagates reliable labels using geometric feature propagation, enabling efficient online adaptation to new environments.
Shin et al. \cite{9879729} enable rapid model adaptation across domains (e.g., different sensors, synthetic-to-real, day-to-night) without source data through intra-modal pseudo-label optimization (Intra-PG) and inter-modal pseudo-label refinement (Inter-PR).
HGL \cite{zou2024hgl} enhances model generalization across different scenarios through geometric learning at three levels: local (point-level), global (object-level), and temporal (frame-level).
\textcolor{black}{Most recently, Wang et al. \cite{WANG2025422} propose a TTA method tailored for addressing domain shift in geospatial point cloud semantic segmentation. It adapts to new domains by dynamically updating batch normalization statistics and improves classification accuracy through information maximization and reliability-constrained pseudo-labeling. However, this method primarily focuses on the TTA setting involving a single, static target domain, and lacks mechanisms to counter the catastrophic forgetting and error accumulation inherent in continual adaptation scenarios.
Unlike Wang et al., our APCoTTA incorporates specifically designed modules to mitigate the issues of catastrophic forgetting and error accumulation that arise during the model's long-term adaptation to unlabeled target domains.}

\subsection{Continual Test-Time Adaptation}

Continual test-time adaptation (CTTA) refers to the process where a model continuously adapts to a stream of data from a dynamic environment during inference to handle distribution shifts while maintaining performance on previously encountered tasks.
It has gained widespread attention in the fields of image classification \cite{maharana2025palm,tian2024parameter} and segmentation \cite{Wang2022ContinualTD,Wang2023ContinualTD,zhu2024reshaping}.

CoTTA \cite{Wang2022ContinualTD}, a pioneering CTTA method, builds on self-training with a weight-averaged teacher and augmented mean predictions to improve pseudo-label quality and reduce error accumulation. It also restores partial network weights to source parameters to preserve knowledge and prevent forgetting.
Zhu et al. \cite{zhu2024reshaping} dynamically store high-confidence samples for replay training to reduce error accumulation. Additionally, they maintain inter-class topology using a class relationship graph (CRG) to prevent catastrophic forgetting. PALM \cite{maharana2025palm} introduces a pseudo-label-free adaptive learning rate strategy that uses the gradient norm of the KL divergence to measure prediction uncertainty and select trainable layers while freezing the rest. It then dynamically adjusts layer-wise learning rates based on parameter sensitivity to support adaptation in continuously changing environments. Inspired by this work, we use gradient norms to identify and freeze stable layers. Unlike PALM, which adjusts layer-wise learning rates, our design is specifically tailored for airborne point clouds and mitigates error accumulation and catastrophic forgetting through entropy and parameter weighting rather than learning rate modulation.

However, research on CTTA for point clouds remains relatively underexplored. Cao et al. \cite{Cao2023MultiModalCT} propose CoMAC, a method designed for real-world scenarios with continually changing target domains in multimodal semantic segmentation. It employs a two-stage modality-aware pseudo-labeling mechanism to adaptively select the more reliable modality and suppress noise. Additionally, it constructs class-wise momentum queues to dynamically update class centers and incorporates pseudo-source feature replay to mitigate catastrophic forgetting. PCoTTA \cite{jiang2024pcotta} prevents catastrophic forgetting by balancing the distance between source prototypes and learnable prototypes. It also employs a Gaussian distribution to align target data features with source domain features, reducing error accumulation. Although these studies have explored CTTA for point clouds, research on ALS point cloud semantic segmentation remains lacking.

\begin{figure*}[!htbp]
  \centering
\includegraphics[width=\textwidth, height=9.5cm, keepaspectratio]{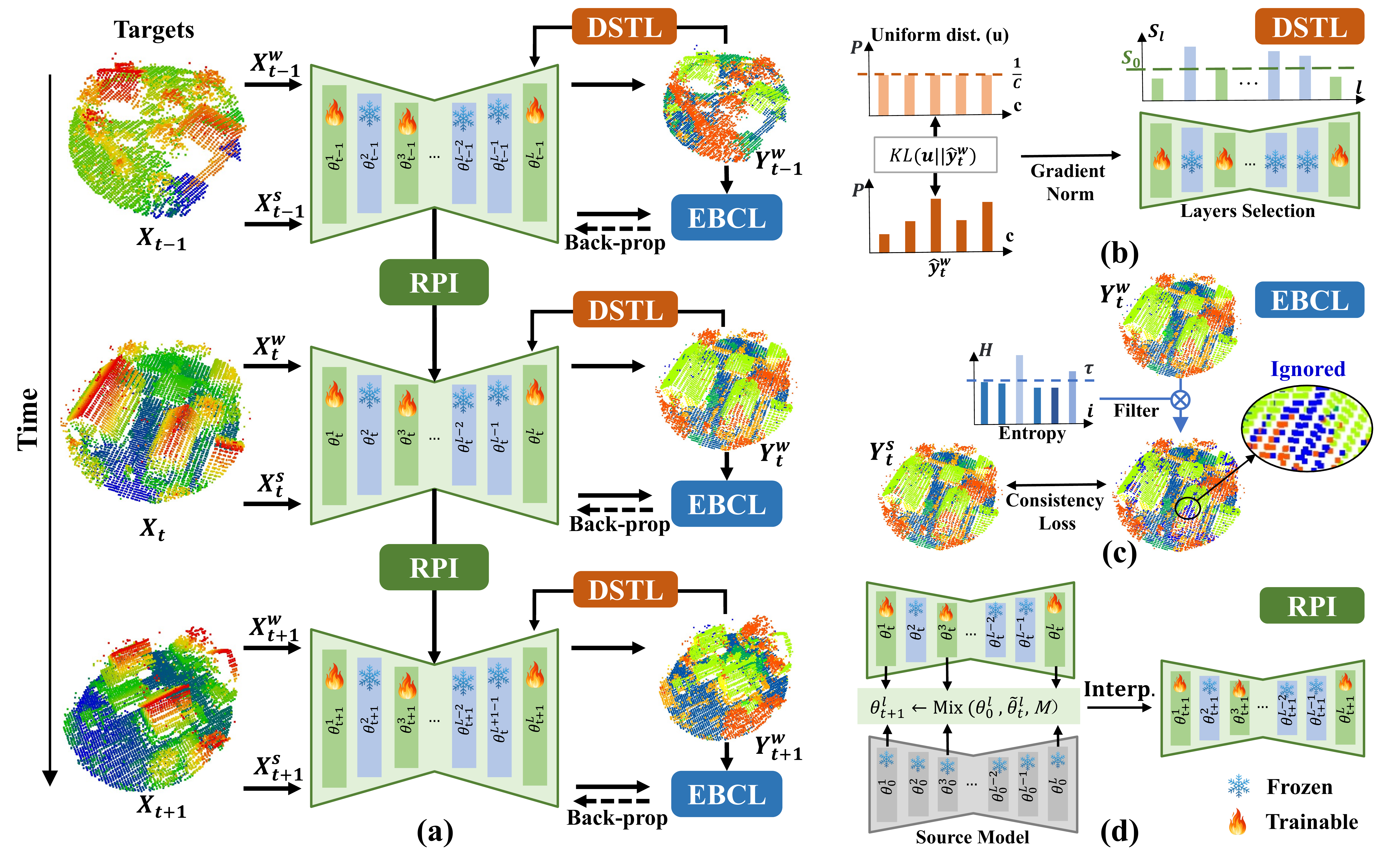}
  \caption{ 
    \textcolor{black}{Overview of the proposed framework}. \textbf{(a)} The overall pipeline at time step $t$. The target domain batch $X_{t}$ is processed with both weak ($X_t^w$) and strong ($X_t^s$) augmentations and fed into the model $f_{\theta_{t}}$. The ``Fire'' and ``Snowflake'' icons explicitly indicate the trainable and frozen status of layers. \textbf{(b)} The Dynamic Selection of Trainable Layers (DSTL) module calculates the gradient norm based on the KL divergence to generate the selection mask $S_l$, identifying low-confidence layers for updates. \textbf{(c)} The Entropy-Based Consistency Loss (EBCL) module utilizes entropy statistics to filter and output high-confidence samples for reducing error accumulation. \textbf{(d)} The Randomized Parameter Interpolation (RPI) module randomly selects a subset of trainable parameters and blends them with their pretrained counterparts (Source Model) to further alleviate catastrophic forgetting.
  }
  \label{fig:framework}
\end{figure*}

\subsection{Corruptions in 3D Semantic Segmentation}

CTTA offers an effective solution to the challenge of continuously evolving domain shifts, but its performance relies on datasets that accurately capture real-world environmental variations.
Ideally, collecting data under diverse weather conditions would provide the highly realistic corruptions; however, this approach is costly and limited in scope. A more cost-effective alternative is synthesizing various corruption types on clean datasets.
ImageNet-C \cite{hendrycks2019benchmarkingneuralnetworkrobustness} pioneers the introduction of various types of image distortions in image classification. 
Similarly, corrupted datasets have also emerged in fields such as image segmentation \cite{kamann2021benchmarking} and object detection \cite{michaelis2020benchmarkingrobustnessobjectdetection}.
For 3D point clouds, research has already established corruption benchmarks for tasks such as point cloud classification \cite{ren2022benchmarking}, object detection \cite{Dong2023BenchmarkingRO,Kong2023Robo3DTR}, and semantic segmentation \cite{Kong2023Robo3DTR,yan2024benchmarking}.
However, they primarily focus on indoor or autonomous driving scenarios. To address this gap, we similarly simulate corruption types on ALS point cloud datasets such as ISPRS \cite{rottensteiner2012isprs} and H3D \cite{kolle2021hessigheim}, providing a benchmark for CTTA research in ALS point clouds.

\section{Method}\label{Third}

\subsection{Preliminaries and Overall Framework}\label{Third_5}

Continual test-time adaptation aims to adapt a source model  $f_{\theta_0}$ pretrained on a labeled source domain $(\mathcal{X}^{S},\mathcal{Y}^{S})$ with $C$ semantic categories to a sequence of evolving,
unlabeled target domains such as $\{ \mathcal{D}^T_1, \mathcal{D}^T_2, \dots, \mathcal{D}^T_n \}$.
During this process, the source data remain inaccessible due to privacy and storage constraints, and the model has no knowledge of when domain shifts occur. Formally, at each time step $t$, the current model $f_{\theta_t}$ receives a batch of unlabeled target-domain inputs $X_t \in \mathbb{R}^{B \times N \times D_{in}}$ (where $B$ is the batch size, $N$ is the number of points, and $D_{in}$ denotes the input feature dimension), makes predictions, and updates itself to \(f_{\theta_{t+1}}\).
To enable robust adaptation, we adopt a teacher-student consistency strategy. Specifically, for each unlabeled target batch $X_t$, we generate two augmented views: a weakly augmented view denoted as $X_t^w$ (e.g., applying mild rigid transformations) and a strongly augmented view denoted as $X_t^s$ (e.g., applying aggressive jittering).
At time step \( t \), the complete model parameters \( \theta_t \) consist of a set of parameter matrices, expressed as
\(\theta_t = \{\theta_t^1, \theta_t^2, \dots, \theta_t^L\}\), 
where \( \theta_t^l \) denotes the parameters of the \( l \)-th trainable layer at time step \( t \), and \( L \) denotes the total number of all trainable layers.
\textcolor{black}{Unlike 2D images, ALS point clouds are characterized by irregular sparsity. Furthermore, severe distribution shifts arising from environmental variations during flight acquisition cause standard CTTA frameworks to suffer from severe local overfitting (e.g., to rural scenes), thereby inducing catastrophic forgetting.}

We propose \textbf{APCoTTA}, a continual test-time adaptation framework for airborne LiDAR point cloud semantic segmentation. As illustrated in Fig. \ref{fig:framework}, APCoTTA consists of three modules: Dynamic Selection of Trainable Layers (DSTL), Entropy-Based Consistency Loss (EBCL), and Randomized Parameter Interpolation (RPI). We adopt a shared-weight framework where BatchNorm layers use the mean and variance of the current batch, and each input is augmented with both strong and weak perturbations. In the DSTL module, we compute the cross-entropy loss against a uniform distribution for each batch and update only the layers whose gradient norms fall below a predefined threshold, keeping the rest frozen. This strategy retains source domain knowledge and mitigates catastrophic forgetting. The EBCL module uses entropy as a confidence measure to filter out low-confidence samples and computes consistency loss only on high-confidence ones, reducing error accumulation. Finally, the RPI module randomly selects a subset of parameters from the trainable layers in each batch and interpolates them with the source domain parameters, balancing source and target knowledge to further alleviate forgetting.

\subsection{Dynamic Selection of Trainable Layers}\label{DSTL}

\textcolor{black}{Compared with regular 2D images, ALS point clouds are irregular and sparse, characterized by heterogeneous local densities and geometric structures. As a result, feature statistics in point cloud networks are inherently less stable than those in image models, making adaptation more susceptible to catastrophic forgetting. This motivates the need for more robust parameter update strategies for point cloud networks during test-time adaptation. Through systematic exploration of parameter update strategies originally developed for image CTTA, we find that commonly used full-parameter update schemes are unsuitable for ALS point clouds, whereas gradient-based partial update methods~\cite{maharana2025palm} are more compatible with the point cloud networks. Based on this observation, we adopt and extend the gradient-driven layer selection mechanism to ALS tasks and develop the Dynamic Selection of Trainable Layers (DSTL) module, which automatically freezes layers that retain robust source domain priors while updating only those sensitive to geometric domain shifts.}

Specifically, at any given inference time step t, we first compute the Kullback-Leibler (KL) divergence \cite{kullback1951information} between the model's softmax predictions on the weakly augmented view, $\hat{y}_t^w = \text{softmax}(f_{\theta_{t}}(X_{t}^{w}))$, and a uniform distribution \textbf{u} = [$\frac{1}{C}$, \ldots, $\frac{1}{C}$] $\in$ $\mathbb{R}^{C}$, where C denotes the total number of classes in the dataset, and then perform backpropagation to obtain the gradients of each parameter.
\begin{equation}\label{eq:initial}
\begin{aligned}
    \mathcal{L}(\theta_t) = \frac{1}{B} \sum_{i=1}^{B} \text{KL}( \text{\textbf{u}} || \hat{y}_{t,i}^w) \\
     \hat{y}_{t,i}^w = \frac{\exp({f_{\theta_{t}}(x_{t,i}^{w})_c / T)}}{\sum_{j=1}^C \exp({f_{\theta_{t}}(x_{t,i}^{w})_j / T)}}
\end{aligned}
\end{equation}
Here, $x_{t,i}^w$ is the $i$-th sample in the batch, \(f_{\theta_{t}}(x_{t,i}^{w})\) represents the logits produced by the model at time step t, and $T$ is the temperature scaling factor (set to 50 following \cite{maharana2025palm}).
The gradient with respect to the parameter \(\theta_{t}\) is:

\begin{equation}\label{eq:initial2}
\begin{aligned}
    \frac{\partial\mathcal{L}(\theta_t)}{\partial\theta_t} = \frac{\partial\text{KL}( \text{\textbf{u}} || \hat{y}_t^w)}{\partial\theta_t}= \frac{1}{C}\sum_{c=1}^{C}\frac{\partial \mathcal{L}_{\text{CE}}(f_{\theta_{t}}(X_{t}^{w}),c)}{\partial \theta_{t}}
\end{aligned}
\end{equation}
This shows that the gradient is equivalent to averaging the derivatives of the cross-entropy loss across all labels.
Thus, we then define the confidence score of the $l$-th layer based on its gradient norm:
\begin{equation}
\label{eq:score}
    S_l(\theta_t) = \frac{1}{K_l}\lVert\frac{\partial\mathcal{L}(\theta_t)}{\partial\theta_t}\rVert_p
\end{equation}
where $\lVert \cdot \rVert_p$ denotes $L_p$-norm, and $K_l$ is the number of parameters in the $l$-th layer, used for normalization. Following \cite{huang2021importance,maharana2025palm}, we adopt the L1 norm in this study. In practice, to mitigate the bias caused by varying layer sizes, we normalize the gradient norm by the number of parameters in each layer.
We set a confidence threshold \(S_{0}\) and, during inference, update only the layers whose scores fall below the threshold, while keeping the other layers frozen.

When processing familiar data, the model tends to make confident predictions, yielding outputs that deviate from a uniform distribution and resulting in higher KL divergence and larger gradient norms during backpropagation. In contrast, unfamiliar data produce outputs closer to a uniform distribution, leading to smaller gradient norms that indicate lower prediction confidence \cite{maharana2025palm}. We define layers with gradient norms below a predefined threshold \(S_{0}\) as low-confidence layers. By updating only these layers while freezing the high-confidence ones, the model improves prediction confidence and preserves source domain knowledge, thereby mitigating catastrophic forgetting.

\subsection{Entropy-Based Consistency Loss}\label{EBCL}

We input strong augmentation data \(X_{t}^{s}\) and weak augmentation data \(X_{t}^{w}\) separately into the same network $f_{\theta_t}$. Following \cite{Wang2022ContinualTD,maharana2025palm}, we regularize the model by enforcing a consistency loss between the two predictions.
\textcolor{black}{ However, error accumulation remains a significant risk due to the inherent physical noise in ALS data, such as solar background noise caused by intense sunlight, which leads to low-confidence predictions. Standard consistency losses indiscriminately align predictions across all samples \cite{Wang2022ContinualTD}, which can inadvertently propagate erroneous pseudo-labels from these low-confidence points.}
Given the success of \cite{zhu2024reshaping,WANG2025422} in leveraging Shannon entropy \cite{shannon1948mathematical} to select high-confidence samples, we introduce an entropy-based filtering mechanism to discard samples with extremely low confidence.
\begin{equation}
\label{eq:entropy}
H(\hat{y}_{t,i}^w) = -\sum_{c=1}^{C} \hat{y}_{t,i,c}^w \log \hat{y}_{t,i,c}^w
\end{equation}
\textcolor{black}{
Unlike common practices that set thresholds for each class to select high-confidence samples, we employ a fixed global threshold $\tau$ to filter out extremely low-confidence samples. 
We define samples with entropy $H(\hat{y}_{t,i}^w) \ge \tau$ as extremely low-confidence, representing high uncertainty where the model fails to produce a sharp probability distribution. 
Such samples often result from severe sensor noise or ambiguous boundaries common in ALS data. 
The choice of a fixed threshold over a dynamic one is crucial for stability. 
In ALS scenarios, batch statistics can fluctuate significantly due to local environmental factors (e.g., reflection saturation). 
A fixed $\tau$ provides a consistent standard, ensuring that the consistency loss is computed only on structurally reliable samples, thus mitigating unstable gradients caused by batch-level outliers. 
Finally, we use the remaining samples for consistency loss:}

\begin{equation}
\label{eq:consistency}
\mathcal{L}_{\text{con}} = \frac{1}{N_{rel}} \sum_{i=1}^{B} \mathbb{I}(H(\hat{y}_{t,i}^w) < \tau) \cdot \left( - \sum_{c=1}^{C} \hat{y}_{t,i,c}^w \log \hat{y}_{t,i,c}^s \right)
\end{equation}
where \( \tau \) is the entropy threshold, $\mathbb{I}(\cdot)$ is the indicator function, and $N_{rel}$ is the number of reliable samples (to normalize the loss).
By filtering out low-confidence samples and retaining more accurate ones, we ensure gradient stability and mitigate error accumulation.

\subsection{Randomized Parameter Interpolation}\label{RPI}
Through the DSTL and EBCL modules, we freeze most layers to retain source domain knowledge, mitigate catastrophic forgetting, and selectively backpropagate reliable predictions to reduce error accumulation. Since the target domain is unlabeled, prolonged adaptation may introduce errors and cause drift from the source domain. 
\textcolor{black}{While conventional methods utilize hard stochastic restoration by randomly resetting weights \cite{Wang2022ContinualTD}, such abrupt changes can disrupt the fine geometric calibration required for 3D point clouds. Consequently, we propose RPI as a soft regularization anchor. Instead of implementing hard resets, we randomly select a subset of parameters within the trainable layers identified by the DSTL module and apply weighted regularization to make them converge toward the source domain weights.}
\begin{equation}
M \sim \text{Bernoulli}(p),
\end{equation}
\begin{equation}
\theta_{t+1}^l = M_l \odot \left( \alpha \theta_0^l + (1-\alpha) \tilde{\theta}_{t}^l \right) + (1-M_l) \odot \tilde{\theta}_{t}^l
\end{equation}
Here, $\tilde{\theta}_{t}^l$ denotes the parameters after the gradient update (via $\mathcal{L}_{\text{con}}$), and $\theta_{t+1}^l$ represents the final parameters for the next time step. $M_l$ is a binary mask sampled with probability $p$ (set to 0.01 \cite{Wang2022ContinualTD}), $\alpha$ is a mixing coefficient, and $\odot$ denotes element-wise multiplication.
By weighting a subset of parameters in selected trainable layers with source domain parameters, we retain source knowledge while maintaining adaptability to the target domain, effectively mitigating deviation from the source and mitigating overfitting to the target, thereby mitigating catastrophic forgetting.

\begin{figure}[!htbp]
  \centering
  \includegraphics[scale=0.095]{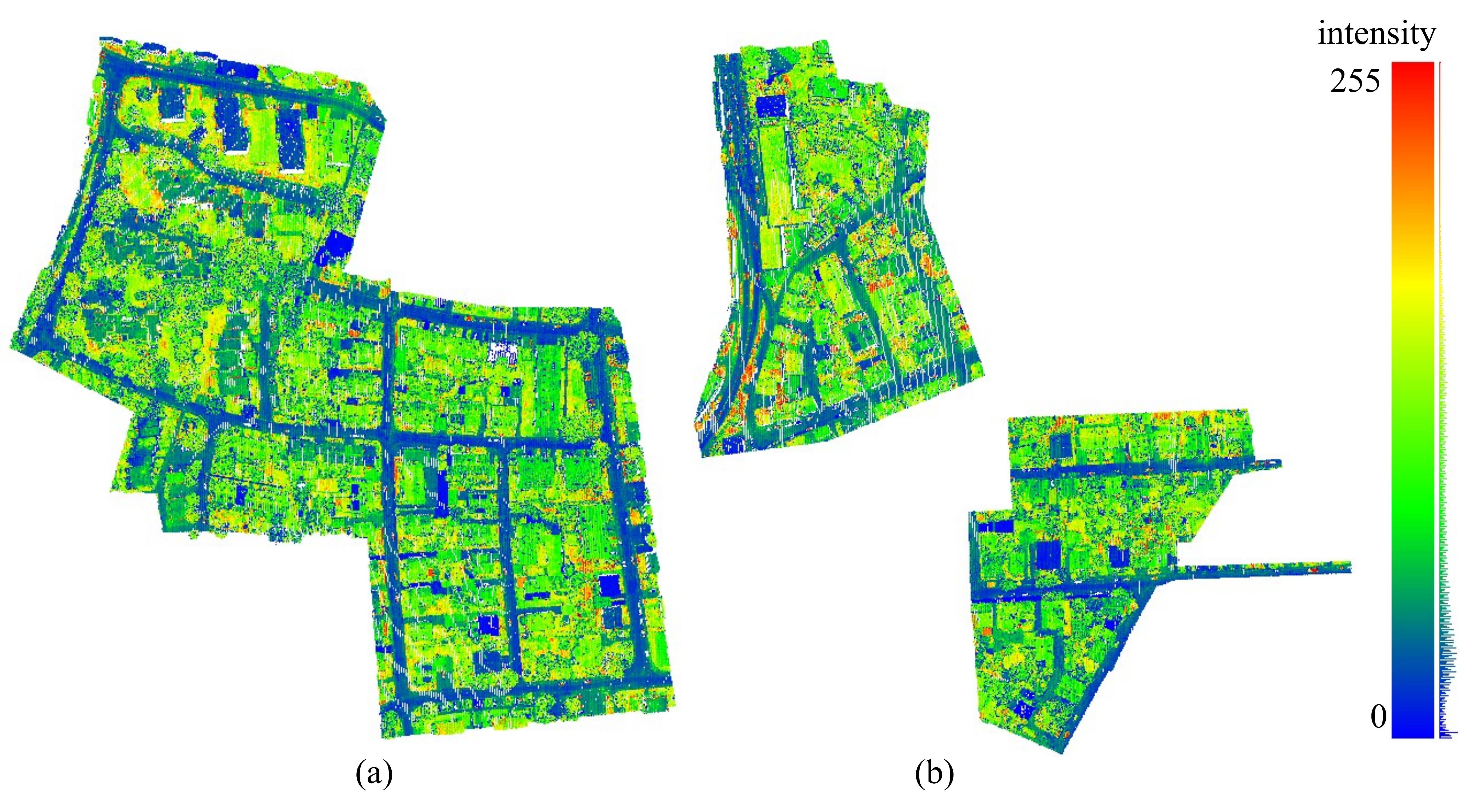}
  \caption{
    The LiDAR intensity map of training (a) and testing (b) sets of ISPRS dataset.
  }
  \label{fig:isprsdataset}
\end{figure}
\begin{figure*}[!htbp]
  \centering
  \includegraphics[scale=0.124]{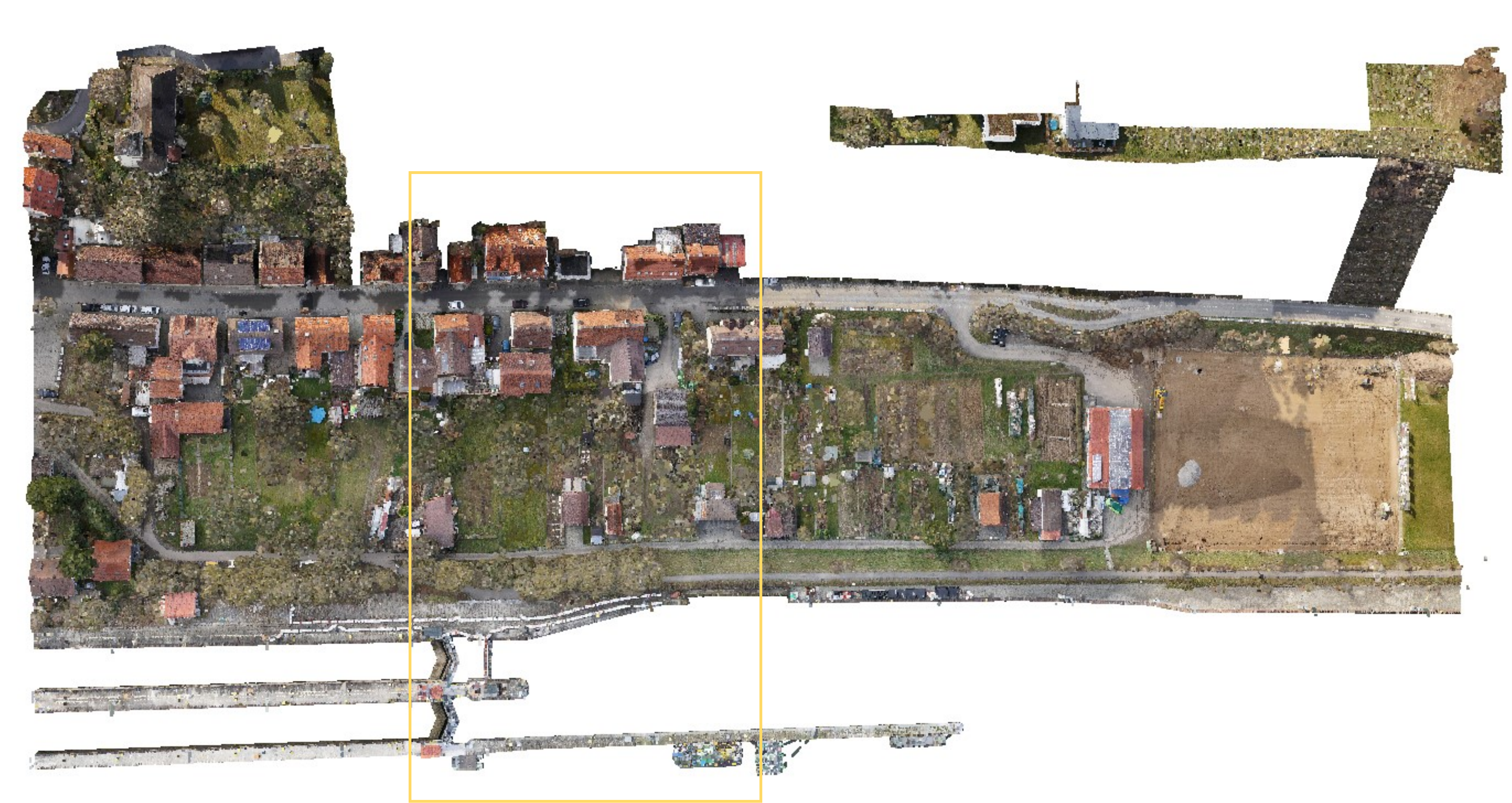}
  \caption{
    The RGB color map of the training and validation sets of the H3D dataset. The validation set is indicated by the yellow box.
  }
  \label{fig:h3ddataset}
\end{figure*}

\section{Experiments}\label{Four}

\subsection{Dataset and Benchmark}\label{datasets}

\subsubsection{Dataset Description}

ISPRS dataset.
The ISPRS Vaihingen 3D Semantic Labeling benchmark (ISPRS) dataset \cite{rottensteiner2012isprs} is collected in the Stuttgart region of Germany using the Leica ALS50 system. The point cloud density ranges from 4 to 7 points/m² and includes nine categories: power lines, low vegetation, impervious surfaces, cars, fences, roofs, facades, shrubs, and trees.
\textcolor{black}{The dataset is explicitly divided into training and testing sets, containing 0.75 million and 0.41 million points, respectively (with an approximate split ratio of 65:35), as shown in Fig. \ref{fig:isprsdataset}.}
The dataset comprises multiple scans with uneven point spacing across the region. To remove redundant points and ensure uniform point density, a subsampling grid size of d = 0.25 m is applied.
The utilized features include {X, Y, Z, Intensity}.

H3D dataset.
The Hessigheim 3D Benchmark (H3D) dataset \cite{kolle2021hessigheim} is collected using a RIEGL VUX1LR scanner mounted on a UAV, acquired through 11 longitudinal and several diagonal scan strips, with an average point density of approximately 800 pts/m². Higher densities occur in some areas due to additional scan strips. The data includes inherent LiDAR attributes such as XYZ coordinates, echo number, and reflectance, with point cloud colors derived via nearest-neighbor interpolation from a textured mesh. 
\textcolor{black}{The training and validation sets contain approximately 59.4 million and 14.5 million points, respectively (with an approximate split ratio of 80:20), as shown in Fig. \ref{fig:h3ddataset}}, covering 11 categories: low vegetation, impervious surfaces, vehicles, urban furniture, roofs, facades, shrubs, trees, soil/gravel, vertical surfaces, and chimneys. Considering the high density of raw data, the data is downsampled to a 0.2 m resolution, utilizing features {X, Y, Z, Reflectance}.

\textcolor{black}{Our selection of these two datasets is driven by their representativeness and complementarity, establishing a rigorous benchmark for CTTA. First, despite its release date, the ISPRS dataset remains a widely recognized ``gold standard'' for ALS segmentation and is actively used in related research, ensuring the continued relevance of our evaluation. \textcolor{black}{Complementarily, the newer H3D (Hessigheim 3D) dataset is widely used in recent research, providing a modern and representative benchmark to further support our evaluation.} Second, the two datasets represent fundamentally different real-world scenarios. ISPRS (Vaihingen) depicts a classic urban environment characterized by complex historic buildings and \textcolor{black}{high-rise residential structures}, acquired by manned aircraft with standard density (4--7 pts/m²). In sharp contrast, H3D captures a distinct village setting, acquired by modern UAV platforms with ultra-high density ($\approx$800 pts/m²). By evaluating on this diverse spectrum, ranging from legacy sparse urban data to modern dense rural data, we verify that APCoTTA is robust across significant variations in sensor platforms, point densities, and semantic environments, which is critical for reliable real-world deployment.
}

\subsubsection{Corruption Robustness Benchmarks}

Following \cite{Dong2023BenchmarkingRO}, we simulate corruptions at the weather and sensor levels, categorizing it into 7 corruption types grouped into five levels.

\textbf{Weather-level corruptions}: 
\textit{Strong Sunlight} interferes with LiDAR signals, and we simulate this corruption by applying strong Gaussian noise to the point clouds. We do not consider extreme weather conditions such as rain, snow, or fog, as UAVs typically do not operate in such environments.

\textbf{Sensor-level corruptions}:
\textit{Density Decrease}: Due to potential non-uniformity in LiDAR sampling density, we simulate this by randomly removing a subset of points.  
\textit{Cutout}: When laser pulses fail to return valid echoes in certain areas, point clouds become incomplete. We model this by discarding points in randomly selected local regions.
Additionally, we address \textbf{LiDAR ranging errors}. To simulate different types of ranging biases, we apply \textit{Gaussian Noise}, \textit{Uniform Noise}, and \textit{Impulse Noise} to the point cloud coordinates.

Finally, we model \textit{Spatial Noise}, which arises from the combined effects of sensor errors, environmental interference, and dynamic factors, resulting in randomly distributed noise points in 3D space. We divide the point cloud into a 3D grid and randomly add noise points within each grid cell to represent these uncertainties.

\begin{figure*}[!t]
  \centering
  \includegraphics[scale=0.16]{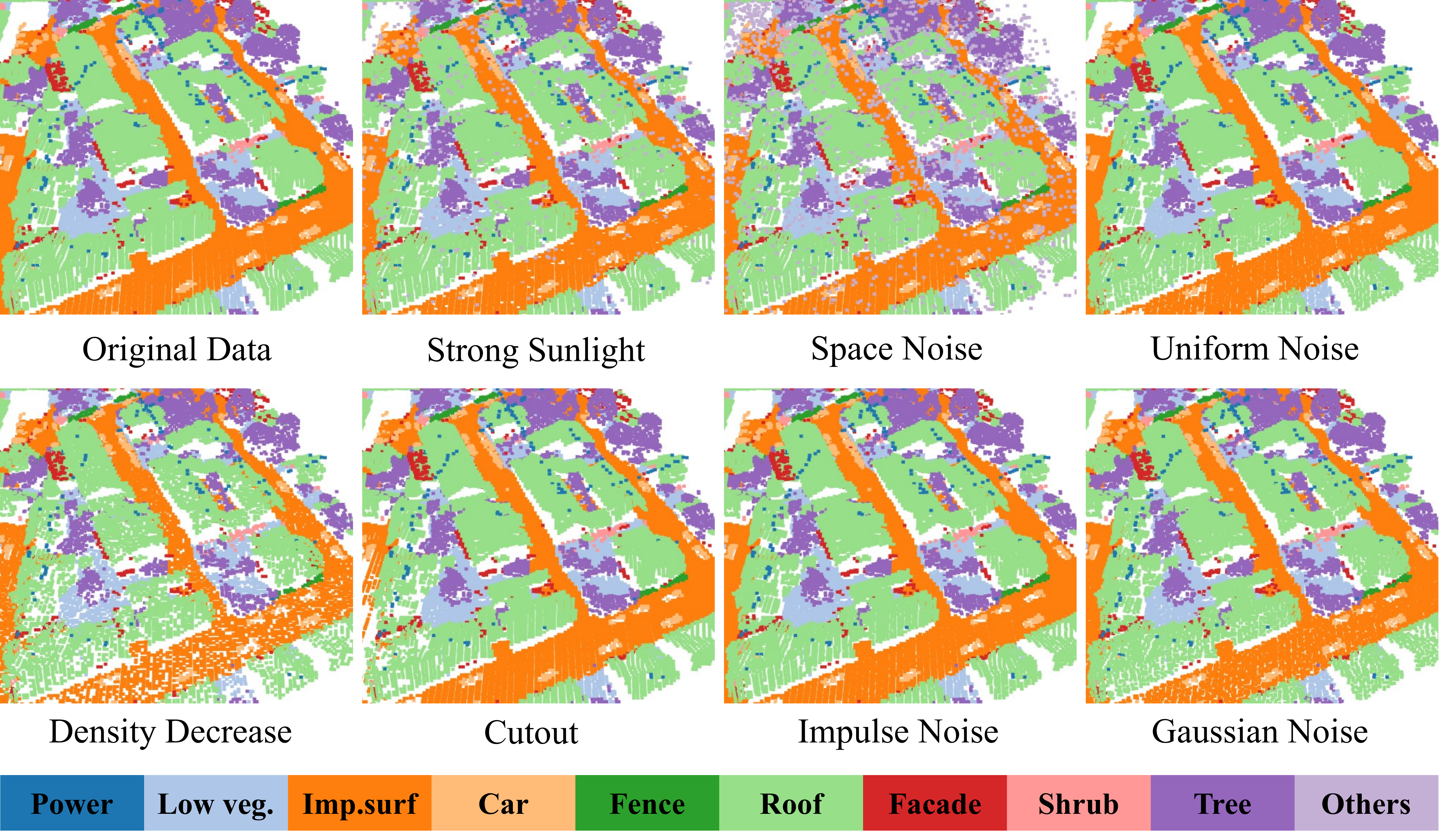}
  \caption{
    Visualization of typical corruption types with the largest corruption severity level 5 in our benchmark ISPRSC dataset.
  }
  \label{fig:isprs_corruption}
\end{figure*}

\begin{figure*}[!t]
  \centering
  \includegraphics[scale=0.16]{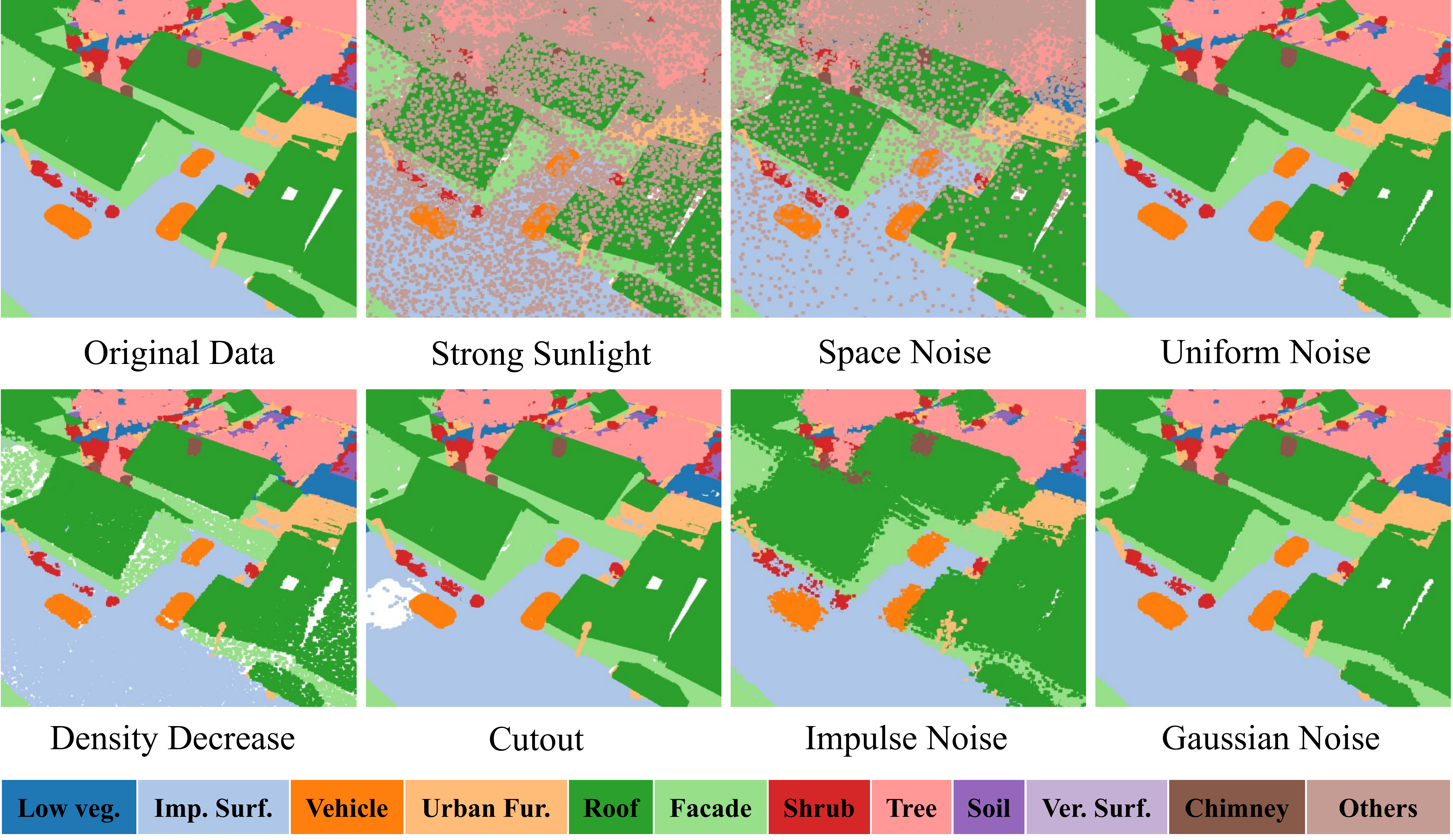}
  \caption{
     Visualization of typical corruption types with the largest corruption severity level 5 in our benchmark H3DC dataset.
  }
  \label{fig:h3d_corruption}
\end{figure*}

By applying the aforementioned disturbance types to the ISPRS 3D test dataset and H3D val dataset, we establish two robustness benchmarks. 
Specifically, we apply the corruptions to their validation or test sets, as shown in Fig. \ref{fig:isprs_corruption},\ref{fig:h3d_corruption}, resulting in ISPRSC and H3DC. More detailed results are provided in the \ref{append:corruption}.


\subsection{Baselines}
We compare our method with multiple CTTA baselines, including Source, BN Stats Adapt \cite{li2016revisitingbatchnormalizationpractical}, Pseudo-label \cite{lee2013pseudo}, TENT-online \cite{Wang2021TentFT}, TENT-continual \cite{Wang2021TentFT}, CoTTA \cite{Wang2022ContinualTD}, PALM \cite{maharana2025palm}, and Wang et al. \cite{WANG2025422}.

For Source, we apply the pretrained model in evaluation mode on the benchmark.
BN Stats Adapt \cite{li2016revisitingbatchnormalizationpractical} predicts using batch normalization statistics from the current batch without modifying network weights.
Pseudo-label \cite{lee2013pseudo} generates pseudo-labels based on the model’s highest-confidence predictions and incorporates them as ground truth for training.
TENT \cite{Wang2021TentFT} minimizes the model’s prediction entropy during testing, updating feature normalization statistics and optimizing channel-wise affine transformation parameters.
TENT-online resets the model to its pretrained state when encountering a new domain.
TENT-continual continuously adapts to new domains without resetting the model.
CoTTA \cite{Wang2022ContinualTD} employs weight averaging and augmentation-averaged pseudo labels while randomly restoring a portion of the weights to the source-pretrained state for online adaptation.
PALM \cite{maharana2025palm} automatically selects layers for adaptation based on model prediction uncertainty and adjusts the learning rate according to parameter sensitivity, enabling continual test-time adaptation.
\textcolor{black}{Wang et al. \cite{WANG2025422} employ a Progressive Batch Normalization (PBN) module that updates BN statistics using the exponential moving average of test batches, while concurrently optimizing the affine parameters of BN layers via a self-supervised loss function.
\textcolor{black}{Specifically dsigned for TTA, this method assumes a static target domain, whereas CTTA addresses ongoing domain shifts.}}

\subsection{Implementation Details}
We evaluate our approach on two continuous test adaptation tasks: ISPRS-to-ISPRSC and H3D-to-H3DC.
\textcolor{black}{Specifically, we first pre-train the source models on the training sets of the original ISPRS and H3D datasets. 
Then, we sequentially adapt the pre-trained network to the corruption benchmarks (ISPRSC and H3DC), which are constructed from the testing set of ISPRS and the validation set of H3D, respectively.
Note that during adaptation, the ground truth labels of the target domain are strictly withheld and used only for performance evaluation.}
Following \cite{Wang2022ContinualTD}, we evaluate all models under the largest corruption severity level 5.
Following \cite{WANG2025422}, We use KPConv \cite{thomas2019kpconv} as the backbone.
To accommodate GPU memory constraints, for the ISPRSC and H3DC datasets, the overlapping spherical sub-cloud radii are 20 m and 11 m, and the batch sizes are 2 and 8. 
All other parameters are kept consistent with the default settings of the KPConv segmentation network. 
We adopt the same Stochastic Gradient Descent (SGD) optimizer with a momentum of 0.98 and an initial learning rate of \(10^{-2}\). 
For both datasets, the threshold $S_0$ for the DSTL module, $\tau$ for the EBCL module, and $\alpha$ for the RPI module are set to 0.001, 0.8, and 0.999, respectively. 
All models are implemented in PyTorch 1.8.2 and trained on a single GeForce GTX 3090 24GB GPU.

\begin{figure*}[!htbp]
  \centering
  \includegraphics[scale=0.09]{image/isprs_result.pdf}
  \caption{
    \textcolor{black}{Local qualitative comparison of semantic segmentation on the ISPRS to ISPRSC CTTA task.}
  }
  \label{fig:isprs_result1}
\end{figure*}

\begin{table*}[!htbp]
\belowrulesep=0pt
\aboverulesep=0pt
\normalsize
\caption{\textcolor{black}{Semantic segmentation results (mIoU in \%) on the ISPRS-to-ISPRSC CTTA task. All results are evaluated with the largest corruption severity level 5 in an online manner. Bold text indicates the best.}}
\label{tab:isprs-c}
\centering
\resizebox{0.75\textwidth}{!}{
\begin{tabular}{l|ccccccc|cc} 
\toprule
Time & \multicolumn{7}{c|}{$t\xrightarrow{\makebox[\dimexpr 20\width][c]{\quad}}$} & ~ \\
\midrule
\multirow{4}{*}{Method} & \multirow{4}*{\rotatebox{75}{\textcolor{black}{sunlight}}} & \multirow{4}*{\rotatebox{75}{space}} & \multirow{4}*{\rotatebox{75}{uniform}} & \multirow{4}*{\rotatebox{75}{density}} & \multirow{4}*{\rotatebox{75}{cutout}} & \multirow{4}*{\rotatebox{75}{impulse}} & \multirow{4}*{\rotatebox{75}{gaussian}} & \multirow{4}*{{Mean mIoU}}  & \multirow{4}*{{Mean OA}} \\
~ & ~ & ~ & ~ & ~ & ~ & ~ & ~ & ~ & ~ \\
~ & ~ & ~ & ~ & ~ & ~ & ~ & ~ & ~ & ~ \\
~ & ~ & ~ & ~ & ~ & ~ & ~ & ~ & ~ & ~ \\
\midrule
Source & 34.98 & 38.57 & 35.95 & 49.60 & 50.02 & 48.72 & 27.91 & 40.82 & 68.15 \\
BN Stats Adapt \cite{li2016revisitingbatchnormalizationpractical} & 39.38 & 39.20 & 48.52 & 49.54 & 48.20 & 51.20 & 40.05 & 45.15 &72.89\\
Pseudo-label~\cite{lee2013pseudo} & \textbf{40.03} & 39.40 & 37.91 & 35.85 & 34.87 & 36.66 & 29.87 & 36.65 &73.49\\
TENT-online \cite{Wang2021TentFT} & 38.75 & 43.00 & 50.05 & 49.57 & 48.83 & 51.90 & 42.41 & 46.36 &74.50\\
TENT-continual \cite{Wang2021TentFT} & 39.14 & 42.13 & 50.46 & 49.21 & 48.56 & 50.92 & 40.28 & 45.81 &75.07\\
CoTTA \cite{Wang2022ContinualTD} & 38.52 & 38.47 & 49.42 & 48.86 & 50.18 & 50.91 & 42.69 & 45.58 &73.10 \\
PALM \cite{maharana2025palm} & 38.35 & 40.82 & 51.79 & \textcolor{black}{49.97} & 50.34 & 52.98 & 38.87 & 46.16 &73.88\\
Wang et al. \cite{WANG2025422} & 39.97 & \textbf{47.85} & 50.32 & 49.35 & 48.18 & 50.83 & 43.93 & 47.21 & 75.93 \\
APCoTTA (Ours) & 39.84 & 45.12 & \textbf{52.83} & \textbf{53.31} & \textbf{51.89} & \textbf{53.96} & \textbf{51.23}  & \textbf{49.74}  & \textbf{77.43} \\
\bottomrule
\end{tabular}
}
\end{table*}

\begin{figure*}[!htbp]
  \centering
  \includegraphics[scale=0.09]{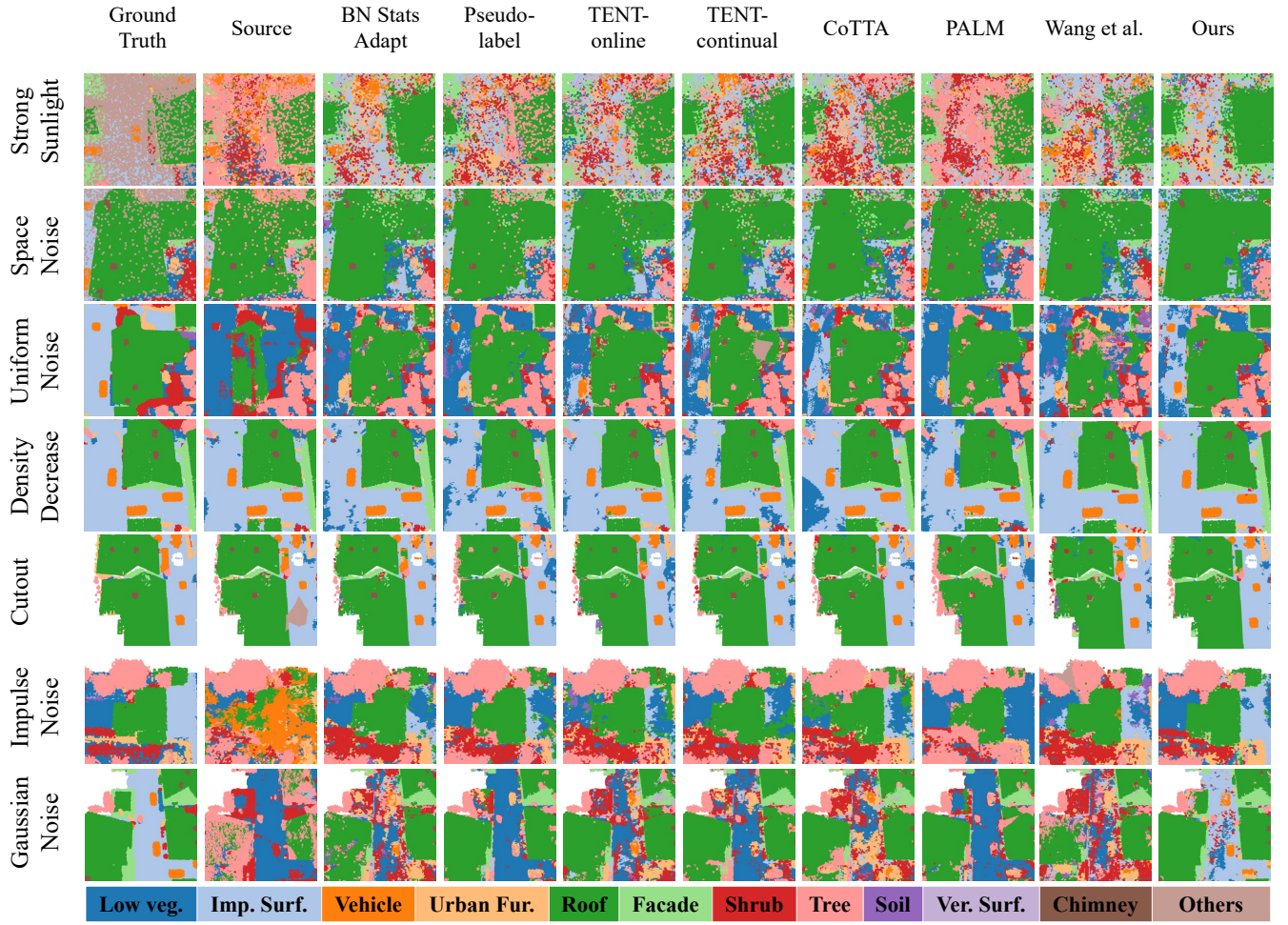}
  \caption{
     \textcolor{black}{Local qualitative comparison of semantic segmentation on the H3D to H3DC CTTA task.}
  }
  \label{fig:h3d_result1}
\end{figure*}

\begin{table*}[!htbp]
\belowrulesep=0pt
\aboverulesep=0pt
\normalsize
\caption{\textcolor{black}{Semantic segmentation results (mIoU in \%) on the H3D-to-H3DC CTTA task. All results are evaluated with the largest corruption severity level 5 in an online manner. Bold text indicates the best.}}
\label{tab:h3d-c}
\centering
\resizebox{0.75\textwidth}{!}{
\begin{tabular}{l|ccccccc|cc} 
\toprule
Time & \multicolumn{7}{c|}{$t\xrightarrow{\makebox[\dimexpr 20\width][c]{\quad}}$} & ~ \\
\midrule
\multirow{4}{*}{Method} & \multirow{4}*{\rotatebox{75}{\textcolor{black}{sunlight}}} & \multirow{4}*{\rotatebox{75}{space}} & \multirow{4}*{\rotatebox{75}{uniform}} & \multirow{4}*{\rotatebox{75}{density}} & \multirow{4}*{\rotatebox{75}{cutout}} & \multirow{4}*{\rotatebox{75}{impulse}} & \multirow{4}*{\rotatebox{75}{gaussian}} & \multirow{4}*{{Mean mIoU}}  & \multirow{4}*{{Mean OA}} \\
~ & ~ & ~ & ~ & ~ & ~ & ~ & ~ & ~ & ~\\
~ & ~ & ~ & ~ & ~ & ~ & ~ & ~ & ~ & ~ \\
~ & ~ & ~ & ~ & ~ & ~ & ~ & ~ & ~ & ~ \\
\midrule
Source & 23.18 & 39.41 & 18.70 & 50.77 & \textbf{59.69} & 21.52 & 13.62 & 32.41 & 65.40\\
BN Stats Adapt \cite{li2016revisitingbatchnormalizationpractical} & 31.16 & 46.06 & 37.28 & 52.17 & 57.29 & 30.26 & 32.64 & 40.98 &71.89\\
Pseudo-label~\cite{lee2013pseudo} & 33.27 & 45.59 & 37.36 & 50.37 & 54.81 & 28.80 & 30.88 & 40.15 &72.74\\
TENT-online \cite{Wang2021TentFT} & 30.86 & 46.76 & 37.05 & \textbf{53.22} & 57.56 & 28.02 & 32.83 & 40.90 &71.71\\
TENT-continual \cite{Wang2021TentFT} & 30.48 & 47.26 & 37.47 & 51.29 & 57.03 & 28.23 & 30.70 & 40.35 &72.01\\
CoTTA \cite{Wang2022ContinualTD} & 30.39 & 46.41 & 37.54 & 53.17 & 56.63 & 28.55 & 31.73 & 40.63  &72.01\\
PALM \cite{maharana2025palm} & 36.35 & \textbf{49.60} & 37.13 & 50.74 & 54.92 & 25.53 & 32.22 & 40.93 &72.65\\
Wang et al. \cite{WANG2025422} & 31.19 & 46.14 & 37.58 & 53.12 & 57.68 & 32.00 & 31.52 & 41.32 & 70.36 \\
APCoTTA (Ours) & \textbf{38.42} & 48.19 & \textbf{46.00} & 53.17 & 57.19 & \textbf{36.51} & \textbf{44.07}  & \textbf{46.22}  & \textbf{77.25}\\
\bottomrule
\end{tabular}
}
\end{table*}

\subsection{Evaluation Metrics}
We evaluate the performance of our method using overall accuracy (OA) and mean intersection over union (mIoU).
Overall accuracy represents the proportion of data correctly classified on the test dataset.
Mean intersection over union represents the average intersection over union for each class on the test dataset.
The expression is as follows:

\begin{equation}
OA = \frac{\sum_{i=1}^{C} TP_i}{N}
\end{equation}

\begin{equation}
mIoU = \frac{1}{C} \sum_{i=1}^{C} \frac{TP_i}{TP_i + FP_i + FN_i}
\end{equation}
where \(N\) represents the total number of samples across all classes, $C$ denotes the total number of classes, and $TP_i$, $FP_i$, and $FN_i$ represent the number of true positives, false positives, and false negatives for class $i$ in a confusion matrix, respectively.

\subsection{Experiments and Analyses}\label{experiment}

\subsubsection{ISPRS to ISPRSC}

To validate the effectiveness of our method, we compare it with several existing approaches on the continual test-time semantic segmentation task from ISPRS to ISPRSC. As shown in Tab. \ref{tab:isprs-c}, directly using the non-adapted pretrained model results in a low mIoU of 40.82\%, mainly due to the significant domain shift between source and target domains. The global mean and variance from the source domain fail to generalize to the target domain, and the performance is further degraded by catastrophic forgetting and error accumulation.
The \textcolor{black}{BN Stats Adapt \cite{li2016revisitingbatchnormalizationpractical}} method, which updates only the batch-wise mean and variance while keeping other parameters fixed, improves performance by 4.33\% in mIoU. In contrast, pseudo-labeling methods lead to mIoU drops, as incorrect pseudo-labels introduce cumulative errors and misleading signals, reducing the model’s adaptability.
\textcolor{black}{TENT-online\cite{Wang2021TentFT}} leverages prior knowledge of domain shifts, resetting the model to its pretrained state upon domain change, which further enhances adaptation. However, this assumption is impractical, as domain change information is typically unavailable in real scenarios. \textcolor{black}{TENT-continual \cite{Wang2021TentFT}} adapts continuously without resetting the model and achieves an improvement of about 5\% in mIoU, though it still underperforms compared to the online version, indicating persistent issues with forgetting and error accumulation.
\textcolor{black}{CoTTA \cite{Wang2022ContinualTD}} adopts a Mean Teacher model to refine pseudo-labels and periodically resets model weights, achieving significant gains. \textcolor{black}{PALM \cite{maharana2025palm}} preserves most of the pretrained knowledge and adjusts the learning rate dynamically, resulting in a 5.34\% improvement in mIoU.
\textcolor{black}{Wang et al. \cite{WANG2025422} propose a TTA method tailored for geospatial point clouds, achieving a superior mIoU of 47.21\%, surpassing previous baselines. However, it still remains inferior to our APCoTTA (49.74\%), as its standard TTA design lacks specific mechanisms to handle the continuous domain shifts inherent in the CTTA setting.}
Compared to all these methods, our approach achieves the best performance across multiple target domains, with an mIoU improvement of approximately 9\%, as highlighted in bold in Tab. \ref{tab:isprs-c}. By filtering out low-confidence samples, we effectively mitigate error accumulation. Additionally, we randomly select a subset of parameters to be weighted with the pretrained model, balancing target adaptation and the retention of source knowledge, thereby reducing catastrophic forgetting.

To further demonstrate the effectiveness of our method, we visualize several local examples in Fig. \ref{fig:isprs_result1} and provide global results in the \ref{append:result}. 
Compared with other methods, our approach demonstrates higher accuracy and robustness in recognizing different classes in the target domain, while others commonly suffer from severe class confusion.

\begin{figure}[!htbp]
  \centering
  \includegraphics[scale=0.086]{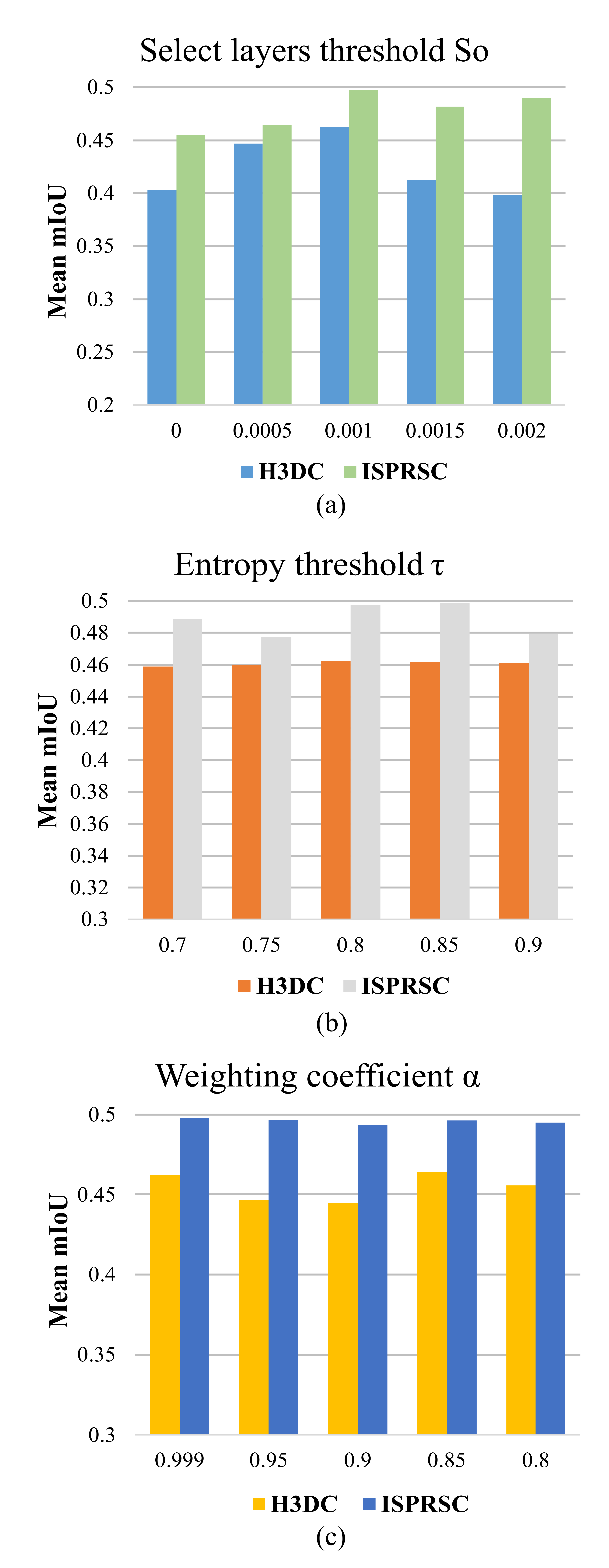}
  \caption{\textcolor{black}{Hyperparameter sensitivity analysis. }}
  \label{fig:parameter}
\end{figure}

\subsubsection{H3D to H3DC}

To further demonstrate the effectiveness of our approach, we evaluated it on a more challenging continual test-time adaptation task, H3D to H3DC. As shown in Tab. \ref{tab:h3d-c}, our method achieves the best performance among all compared state-of-the-art approaches, with a significant mIoU improvement of 13.8\%. While methods such as \textcolor{black}{BN Stats Adapt \cite{li2016revisitingbatchnormalizationpractical}, Pseudo-label~\cite{lee2013pseudo}, TENT-online\cite{Wang2021TentFT}, TENT-continual \cite{Wang2021TentFT}, CoTTA \cite{Wang2022ContinualTD}, and PALM \cite{maharana2025palm}} achieve around 8\% mIoU improvement, they suffer from severe catastrophic forgetting and error accumulation over prolonged adaptation, leading to poor performance on target domains with impulse and Gaussian noise.
\textcolor{black}{Wang et al. \cite{WANG2025422}, which is designed for remote sensing point clouds, demonstrates strong performance in the early stages and achieves the second-best overall mIoU of 41.32\%. However, its performance degrades noticeably in the later stages of the corruption sequence (e.g., 32.00\% on Impulse and 31.52\% on Gaussian noise), indicating a susceptibility to error accumulation over long-term adaptation. }
In contrast, our method effectively mitigates catastrophic forgetting and error accumulation, resulting in significantly better adaptation in target domains. Additionally, as illustrated in Fig.~\ref{fig:h3d_result1}, our method demonstrates superior class-wise adaptation across multiple categories, whereas other methods exhibit notable class misclassification. More detailed results are provided in the \ref{append:result}.

\subsubsection{Discussion}
\label{sec:dis}
\textcolor{black}{The introduction of the Continual Test-Time Adaptation (CTTA) paradigm into 3D ALS point clouds is not merely a straightforward extension of existing 2D approaches, but is fundamentally driven by the intrinsic characteristics of LiDAR data and real-world deployment constraints. During acquisition, ALS point clouds exhibit continuous and spatially evolving distribution shifts (e.g., transitions from urban to mountainous regions), which are significantly more complex than the typically discrete, appearance-driven shifts in 2D images. As demonstrated in Tabs.\ref{tab:isprs-c} and \ref{tab:h3d-c}, although directly applying image-based CTTA methods (e.g., TENT-continual \cite{Wang2021TentFT} and CoTTA \cite{Wang2022ContinualTD}) yields certain improvements (e.g., an approximately 5\% increase in mIoU, as shown in Tab. \ref{tab:isprs-c}), they still suffer from specific performance bottlenecks in ALS scenarios.}

\textcolor{black}{This performance gap fundamentally stems from the severe discrepancies between 2D images and 3D point clouds in terms of data characteristics and learning mechanisms. Regarding data characteristics, domain shifts in 2D images typically manifest as illumination or style variations on regular pixel grids. In contrast, ALS point clouds are inherently unstructured and sparse, with domain shifts primarily appearing as severe geometric distortions and drastic density fluctuations. Consequently, key assumptions underlying image-based CTTA, such as the premise that data augmentations preserve prediction consistency \cite{Wang2022ContinualTD} are frequently violated in point clouds. In terms of learning mechanisms, image networks operate on regular grids with relatively stable feature statistics. Conversely, point cloud networks process unordered data where neighborhood structures dynamically change with sampling conditions, rendering feature statistics highly unstable. Therefore, directly adopting common image-domain strategies, such as full-parameter updates or naive BatchNorm statistical adjustments \cite{Wang2021TentFT}, inevitably leads to gradient instability, overfitting, and catastrophic forgetting of source-domain geometric knowledge in point cloud scenarios.}

\textcolor{black}{Unlike image-based methods, APCoTTA introduces tailored, domain-specific designs. Considering the complex geometric structures and density variations of airborne point clouds, our DSTL module abandons full-parameter updates. Instead, it utilizes gradient norms as a geometry stability filter to dynamically identify and freeze key layers insensitive to geometric shifts. Furthermore, recognizing that ALS sensors are highly susceptible to capturing unreliable noisy samples, the EBCL module is introduced to explicitly filter low-confidence predictions to mitigate error accumulation. Finally, the RPI strategy is incorporated to smooth model updates. In summary, the significant performance gains achieved by our method, an approximate 9\% improvement in Tab. \ref{tab:isprs-c} and 14\% in Tab. \ref{tab:h3d-c}, are rooted in a deep insight into the geometric and statistical properties of 3D point clouds, rather than a naive transfer of existing image-based CTTA methods.}

\subsection{Ablation Study}

\subsubsection{Effectiveness of Each Component}
\begin{table}[!htbp]
    \centering
\caption{Ablation experiments.}
    \resizebox{0.4\textwidth}{!}{%
    \begin{tabular}{ccccccc}
    \toprule
        \multicolumn{3}{c}{Module} & \multicolumn{2}{c}{ISPRSC} &  \multicolumn{2}{c}{H3DC} \\ \hline
        DSTL & EBCL & RPI & OA & mIoU & OA & mIoU \\ \bottomrule
        ~ & ~ & ~ & 59.21 & 25.83 & 69.00 & 33.37 \\ 
        \checkmark &  ~ &  ~ & 75.17 & 47.51 & 75.51 & 44.25 \\ 
        \checkmark & \checkmark & ~ & 76.81 & 48.81 & 76.90 & 45.58 \\ 
        \checkmark & \checkmark & \checkmark & \pmb{77.43} & \pmb{49.74} & \pmb{77.25} & \pmb{46.22} \\ \hline
    \end{tabular}
    }
    \label{tbl:ablation}
\end{table}
We conduct ablation studies on two continual test-time adaptation tasks, ISPRS to ISPRSC and H3D to H3DC, to evaluate the effectiveness of each module. As shown in Tab. \ref{tbl:ablation}, when all modules are removed, the model adapts to the target domain using only the current batch’s mean and variance, a fully learnable shared-weight network, and a consistency loss. Due to catastrophic forgetting and error accumulation, the model performs poorly, with mIoU of 25.83\% and 33.37\% on the two tasks, respectively.
When the first module, DSTL, is enabled, the model retains most of the useful knowledge from the source domain, which effectively mitigates catastrophic forgetting and significantly improves performance. Enabling the second module, EBCL, further reduces the influence of erroneous information by discarding inaccurate samples, thereby alleviating error accumulation. Finally, when the third module, RPI, is activated, the model maintains strong adaptation to the target domain while preserving alignment with the source domain distribution, further mitigating catastrophic forgetting.

\subsubsection{Hyperparameter Sensitivity}

We analyze the sensitivity of our model to four key hyperparameters in two continuous test-time adaptation tasks (ISPRS to ISPRSC and H3D to H3DC), as shown in Fig. \ref{fig:parameter}:
\textbf{Select layer threshold \(S_{0}\)}: This parameter controls the number of trainable layers. We vary it within the range {0, 0.0005, 0.001, 0.0015, 0.002}. The best performance is achieved around 0.001, with noticeable performance drops when deviating from this value.
\textbf{Entropy threshold \(\tau\)}: This parameter filters out samples with very low confidence. We test values in the range {0.7, 0.75, 0.8, 0.85, 0.9}. The results show that model performance is relatively stable across this range.
\textbf{Weighting coefficient \( \alpha \)}: This parameter regulates the distance between the adaptive model and the source domain. Our method achieves optimal performance when \( \alpha \) is set to 0.999.
\textcolor{black}{\textbf{Sampling probability $p$:} We evaluate $p \in \{0.001, 0.005, 0.01, 0.05, 0.1\}$, and results demonstrate that performance peaks at $p=0.01$. Notably, increasing $p$ to $0.1$ causes a significant drop in mIoU (e.g., $-3.40\%$ on ISPRSC and $-3.12\%$ on H3DC), indicating that excessive restoration overly constrains the model to the source domain, hindering adaptation to extreme geometric shifts. Conversely, lowering $p$ to $0.001$ results in insufficient regularization against forgetting. Thus, $p=0.01$ provides the optimal cumulative regularization effect.}

\section{Conclusions}\label{Five}
In this study, we present the dedicated benchmarks for CTTA in airborne LiDAR point cloud semantic segmentation, filling a critical gap in the field through the release of the ISPRSC and H3DC datasets.
To address the performance degradation caused by continuous domain shifts in real-world scenarios, we propose a CTTA framework specifically designed for airborne LiDAR point clouds, named APCoTTA. It comprises three key components: (1) a dynamic trainable layer selection strategy, which freezes most network layers while updating only a few crucial ones, mitigating catastrophic forgetting and preserving source domain knowledge; (2) an entropy-based consistency loss that filters out low-confidence samples to reduce error accumulation; and (3) a random parameter interpolation mechanism that blends the parameters of trainable layers with their source counterparts to further alleviate forgetting.
Evaluated on our proposed benchmarks, APCoTTA achieves mIoU improvements of approximately 9\% and 14\% over direct inference.

Despite the progress made in this study, several limitations remain. The issue of class imbalance in airborne LiDAR point clouds, which can significantly impact model performance on long-tail or rare classes, has not been thoroughly addressed. Future work may explore more sophisticated re-sampling strategies or class-balanced weighting methods to address this challenge. Furthermore, we plan to extend continuous test-time adaptation to open-set scenarios to handle the emergence of unknown classes during deployment.

\section*{Acknowledgements}

This work was supported by the National Key Research and Development Program of China (2023YFB3907401), and the National Natural Science Foundation of China under Grant 42201481, 42271365.

\bibliographystyle{elsarticle_harv} 
\bibliography{ref}

\clearpage
\appendix
\onecolumn

\subsection{Implementation Details for Corruption Robustness Benchmarks} \label{append:corruption}

\begin{figure*}[!t]
  \centering
  \includegraphics[scale=0.15]{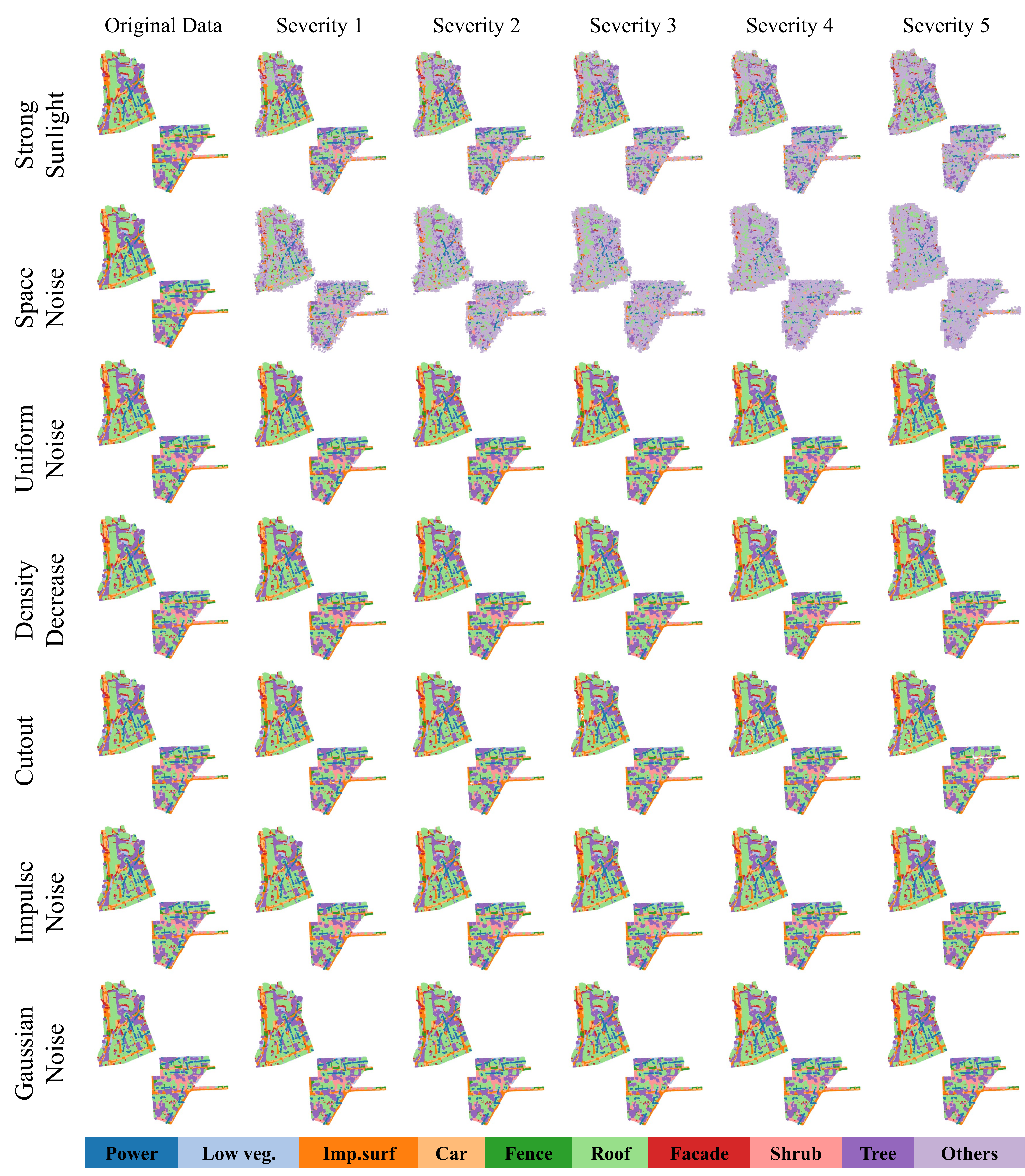}
  \caption{
  Full visualization results for all corruptions in our ISPRSC benchmark.
  }
  \label{fig:isprsOV_corruption}
\end{figure*}

\begin{figure*}[!t]
  \centering
  \includegraphics[scale=0.135]{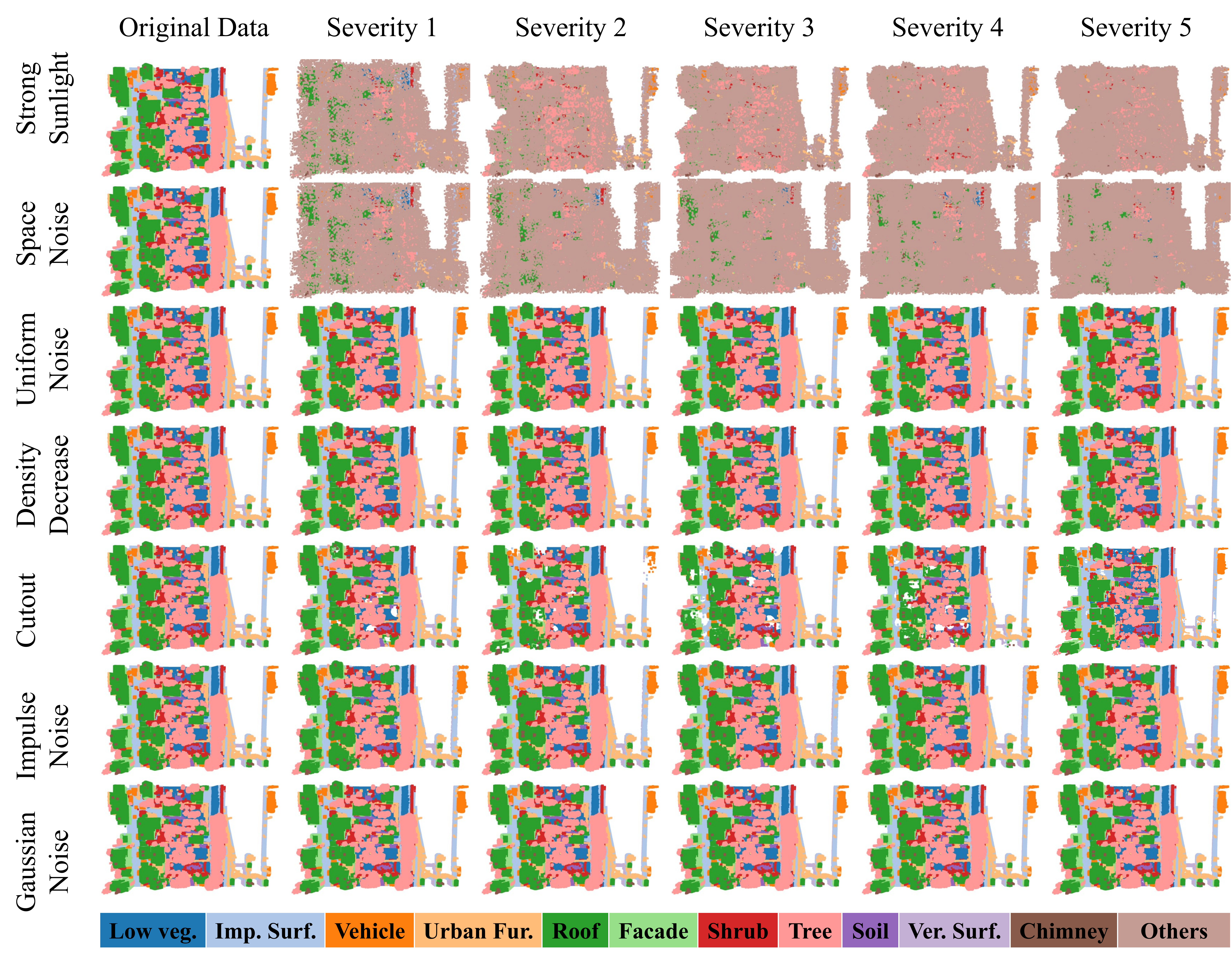}
  \caption{
   Full visualization results for all corruptions in our H3DC benchmark.
  }
  \label{fig:h3dOV_corruption}
\end{figure*}

In this section, we provide the parameters and experimental details for seven common corruptions in the ISPRSC and H3DC benchmarks. Each corruption is divided into five severity levels, with level 5 being the most severe.
Fig \ref{fig:isprsOV_corruption} and Fig \ref{fig:h3dOV_corruption} show all corruption types across all severity levels in the two benchmarks, respectively.

\textbf{\emph{Strong Sunlight.}}
Following the observations in \cite{carballo2020libre,Dong2023BenchmarkingRO}, we simulate this corruption by adding 2m and 1m Gaussian noise to points in the ISPRS and H3D datasets, respectively. The severity levels are defined by noisy point ratios of \{0.7\%, 1.4\%, 2.1\%, 2.8\%, 3.5\%\} for ISPRS and \{0.3\%, 0.6\%, 0.9\%, 1.2\%, 1.5\%\} for H3D.

\textbf{\emph{Density Decrease.}}
Following the observations in \cite{Dong2023BenchmarkingRO}, we randomly remove \{6.02\%, 12.04\%, 18.06\%, 24.08\%, 30.1\%\} of the LiDAR points in ISPRS and \{18.2\%, 36.4\%, 54.6\%, 72.8\%, 91.0\%\} in H3D, respectively.

\textbf{\emph{Cutout.}}
Following the observations in \cite{Dong2023BenchmarkingRO}, we randomly remove \{2, 3, 5, 7, 10\} groups of points for both ISPRS and H3D. Each group contains $\frac{3N}{100}$ points for ISPRS and $\frac{N}{100}$ points for H3D, respectively, and is located within a ball in the Euclidean space, where $N$ denotes the total number of points in the point cloud.

 \textbf{\emph{Gaussian Noise.}}
Following the observations in \cite{Dong2023BenchmarkingRO}, we add Gaussian noise to all points. For the ISPRS dataset, the noise severities are \{0.02002m, 0.04004m, 0.06006m, 0.08008m, 0.1001m\}, and for the H3D dataset, they are \{0.012m, 0.024m, 0.036m, 0.048m, 0.060m\}.

 \textbf{\emph{Uniform Noise.}}
Following the observations in \cite{Dong2023BenchmarkingRO}, we add uniform noise to all points. For ISPRS and H3D datasets, the noise severities are \{0.028m, 0.056m, 0.084m, 0.112m, 0.140m\}.

 \textbf{\emph{Impulse Noise.}}
Following the observations in \cite{Dong2023BenchmarkingRO}, for the ISPRS and H3D datasets, we select the number of points from the sets $\{\frac{11N}{300}, \frac{11N}{250}, \frac{11N}{200}, \frac{11N}{150}, \frac{11N}{100}\}$ and $\{\frac{7N}{300}, \frac{7N}{250}, \frac{7N}{200}, \frac{7N}{150}, \frac{7N}{100}\}$, respectively, to add impulse noise and represent different noise severity levels, where $N$ denotes the total number of points.

 \textbf{\emph{Space Noise.}}
For the ISPRS dataset, the point cloud is divided into 10×10×10 grid cells, and \{5, 10, 15, 20, 25\} noise points are randomly generated per cell based on noise severity, then merged with the original point cloud. For the H3D dataset, the same grid division is applied, with \{100, 200, 300, 400, 500\} noise points generated per cell accordingly.

\subsection{Experimental Results} \label{append:result}

\begin{figure*}[!t]
  \centering
  \includegraphics[scale=0.09]{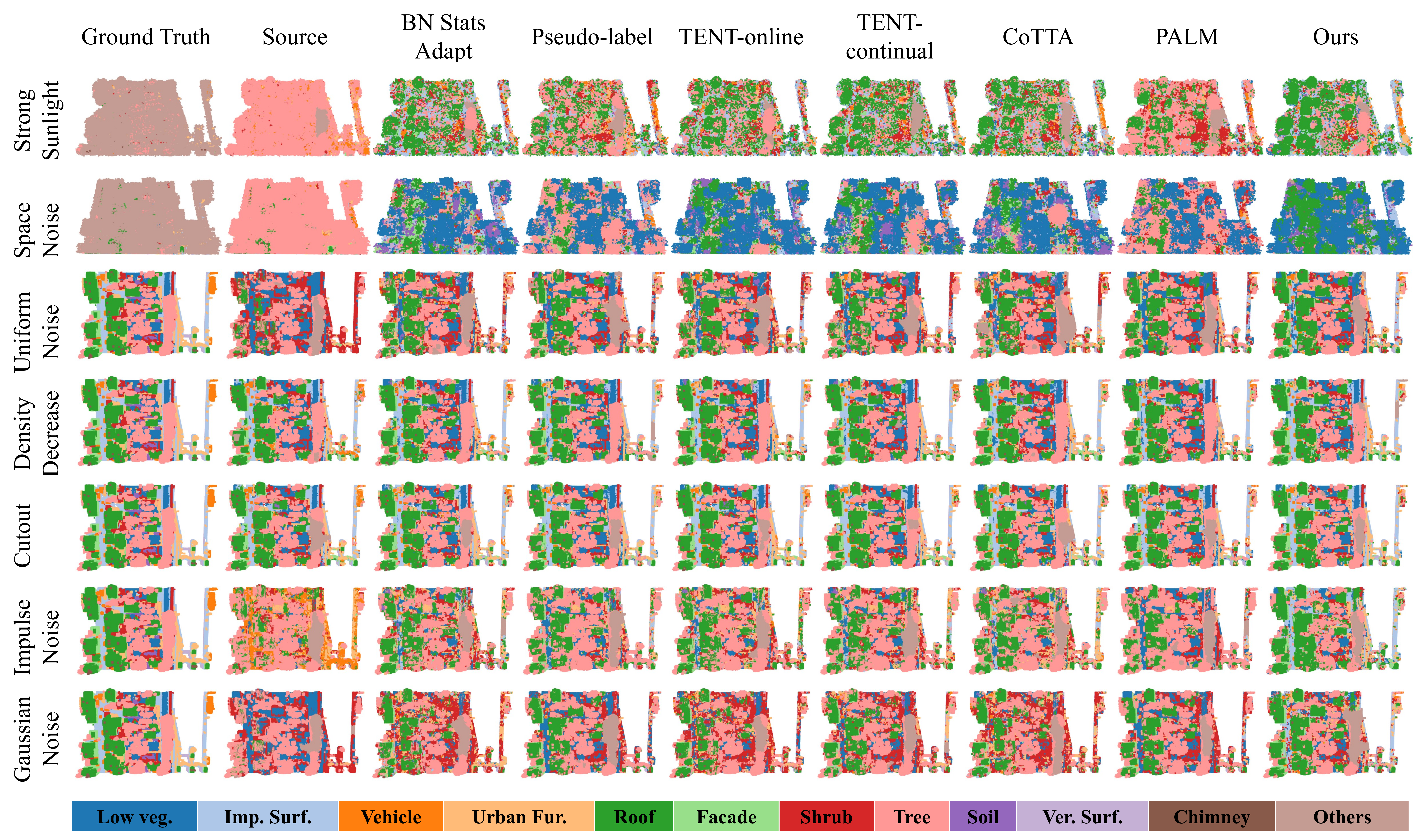}
  \caption{
   \textcolor{black}{Global qualitative comparison of semantic segmentation on the H3D to H3DC CTTA task.}
  }
  \label{fig:h3dOV_result1}
\end{figure*}

In this section, we present global result visualizations on our ISPRSC and H3DC datasets, as shown in Fig \ref{fig:isprsOV_result1} and Fig \ref{fig:h3dOV_result1}.

\begin{figure*}[!t]
  \centering
  \includegraphics[scale=0.09]{image/isprs_overResult.pdf}
  \caption{
    \textcolor{black}{Global qualitative comparison of semantic segmentation on the ISPRS to ISPRSC CTTA task.}
  }
  \label{fig:isprsOV_result1}
\end{figure*}

\end{document}